\documentclass[11pt]{article}
\usepackage[preprint]{acl}
\usepackage{times}
\usepackage{latexsym}
\usepackage[T1]{fontenc}
\usepackage[utf8]{inputenc}
\usepackage{microtype}
\usepackage{inconsolata}
\usepackage{graphicx}
\usepackage{booktabs}
\usepackage{amsmath,amssymb}
\usepackage{xurl}
\usepackage{array}
\usepackage{algorithm}
\usepackage{algpseudocode}
\makeatletter
\newcommand{\StateText}[1]{%
  \State\parbox[t]{\dimexpr\linewidth-\ALG@thistlm\relax}{\raggedright #1\strut}}
\newcommand{\RequireText}[1]{%
  \Require\begingroup
  \settowidth{\@tempdima}{\algorithmicrequire}%
  \parbox[t]{\dimexpr\linewidth-\@tempdima+\labelwidth\relax}{\raggedright #1\strut}%
  \endgroup}
\makeatother
\newcommand{\Owpq}{Output-path-weighted PCA-PQ}
\newcolumntype{L}[1]{>{\raggedright\arraybackslash}p{#1}}

\title{Extreme Compression of Per-Layer Embeddings\\
via Cross-Layer Latent Vector Quantization}
\author{
Jiacheng Liu\textsuperscript{1} \quad
Yaxin Luo\textsuperscript{1} \quad
Xiaohan Zhao\textsuperscript{1} \quad
Peng Sun\textsuperscript{2,3} \\
\bfseries Xinyi Shang\textsuperscript{1,4} \quad
Jiacheng Cui\textsuperscript{1} \quad
Yonina C. Eldar\textsuperscript{5} \quad
Zhiqiang Shen\textsuperscript{1}\thanks{Corresponding author.} \\[4pt]
\textsuperscript{1}Mohamed bin Zayed University of Artificial Intelligence \quad
\textsuperscript{2}Westlake University \\
\textsuperscript{3}Zhejiang University \quad
\textsuperscript{4}University College London \quad
\textsuperscript{5}Weizmann Institute of Science
}
\begin{document}
\maketitle
\begin{abstract}
Scaling embedding capacity alongside Transformer computation is an
emerging trend in language models. This trend makes high compression
rates increasingly useful. Hence, we study extreme compression of Per-Layer
Embeddings (PLE), which follow this trend and account for 35.2\% of the
parameters of Gemma~4 E4B. At such rates, conventional quantizers give each
token--layer vector its own code of only a few bytes. We instead develop OPTIC
(\textbf{O}utput-\textbf{P}ath-fitted, \textbf{T}oken-shared \textbf{I}ndexed
\textbf{C}odes)\footnote{Code: \url{https://github.com/VILA-Lab/optic}}, which stores one quantized code tuple per token and decodes it at
every layer with a layer-specific affine decoder. Codes and decoders are fitted to the PLE output path
rather than to the stored vectors. At about $100\times$ PLE compression of
bfloat16 (BF16) E4B, OPTIC retains an aggregate 97.8\% of the uncompressed
model's performance across seven benchmarks, whereas product quantization, the strongest
of six conventional baselines, retains 66.3\%. At $594\times$ and
$1,002\times$ PLE compression of BF16 E4B, OPTIC retains 90.5\% and 84.1\%.
On the official E4B release with 4-bit linear weights and 16-bit activations
(W4A16), OPTIC retains 96.9\% at about $100\times$ PLE compression. During
inference at about $100\times$ PLE compression, BF16 E4B uses 29.7\% less GPU
memory than with uncompressed PLE at the same key--value cache size, while
running at most 0.40\% slower.
\end{abstract}
\section{Introduction}
Embedding capacity is emerging as a distinct scaling dimension in recent
language-model development. Over-Tokenized Transformers~\citep{huang2025over}
and SCONE~\citep{yu2026scaling} scale multi-gram or contextual $n$-gram
representations; DeepSeek's Engram uses hash-addressed $n$-gram lookup
memory~\citep{cheng2026conditional}; and Qwen3.8-Flash-Next supplements a
125B-parameter language model with 51B parameters of $n$-gram
embeddings~\citep{qiu2026design}. Gemma~4 follows this broader direction with
Per-Layer Embeddings (PLE), first introduced in
Gemma~3n~\citep{google2025gemma3n}, which provide every vocabulary item with a
learned vector at each decoder layer~\citep{team2026gemma}.

We study Gemma~4 E2B and E4B for their strong performance and potential
for practical deployment~\citep{team2026gemma}. PLE accounts for 45.8\% and 35.2\% of the
two checkpoints, respectively, making it a substantial target for
storage reduction as models expand their embedding capacity.

An ordinary input embedding maps a token to one vector at the model's
input. PLE provides a separate vector at each decoder layer: E4B stores
11 million token--layer vectors of width 256, and near $100\times$
compression leaves about five bytes for each. Conventional quantizers spend
these bytes on a separate code for every vector.

OPTIC (Output-Path-fitted, Token-shared Indexed Codes) builds on two
observations. First, PLE stores one vector for every pair of token and
layer, so a compact representation can share storage along either index.
OPTIC shares along both. All tokens of a layer use one layer-specific
affine decoder, whose storage does not grow with the vocabulary size. All
layers of a token use one code tuple, so a token's bytes form one larger
code instead of many small per-layer codes, and the layer decoders preserve
the differences between layers.

Second, the model uses PLE vectors through a gate and an output projection.
The fitting objective should therefore account for reconstruction error
after the gate and output projection, rather than only in the stored
coordinates. OPTIC fits its codes and decoders to minimize error at the
projection output.

We report performance retention across seven benchmarks relative to the
same checkpoint with uncompressed PLE. At about $100\times$ PLE compression, OPTIC
retains 97.8\% on BF16 E4B, whereas the strongest of six conventional
baselines retains 66.3\%. With OPTIC's fitting recipe and nearly equal
storage, independent latent codes for each token--layer vector retain 79.3\%
(Section~\ref{sec:code-sharing}). At $594\times$ and
$1,002\times$ PLE compression,
where about one byte or less remains for each vector, OPTIC retains
90.5\% and 84.1\%. In addition, on the official E4B release with 4-bit
linear weights and 16-bit activations (W4A16), OPTIC retains 96.9\% at about
$100\times$ PLE compression, compared with 82.8\% for the strongest
conventional baseline. On the smaller E2B, OPTIC retains 89.1\%
on BF16 and 91.5\% on the official W4A16 release.

A model compressed with OPTIC also supports inference directly from the compact codes.
On an RTX~4090 with a fixed key--value cache size, compact PLE lowers the
GPU memory of BF16 E4B serving by 29.7\% and adds at most 0.40\% wall time
(Section~\ref{sec:compact-inference}).
\section{Method}
\label{sec:methods}
\subsection{Problem Setup}
For a model with vocabulary size $V$, $L$ PLE-equipped decoder layers, and
PLE width $d$, let $\mathbf E\in\mathbb R^{V\times L\times d}$ denote the
pretrained PLE tensor. Its vector for token
$t\in\{1,\ldots,V\}$ at layer $\ell\in\{1,\ldots,L\}$ is
$\mathbf e_{t\ell}=\mathbf E[t,\ell,:]\in\mathbb R^d$. We call each
$\mathbf e_{t\ell}$ a PLE row. We compress
$\mathbf E$ while keeping the remaining model parameters fixed.

Let
$\boldsymbol\theta$ denote all tensors in a compact representation,
$\operatorname{bytes}(\boldsymbol\theta)$ its resident size, and
$\operatorname{Dec}_{\boldsymbol\theta}(t,\ell)\in\mathbb R^d$ the decoder
output for token $t$ at layer $\ell$. Given a byte budget $B_{\max}$,
the reconstructed row is
\begin{equation}
  \widehat{\mathbf e}_{\boldsymbol\theta,t\ell}
  =\operatorname{Dec}_{\boldsymbol\theta}(t,\ell),
  \qquad
  \operatorname{bytes}(\boldsymbol\theta)\le B_{\max}.
  \label{eq:compact-ple-decoder}
\end{equation}
\subsection{Shared Token Codes}
\begin{figure*}[t]
\centering
\includegraphics[width=\textwidth]{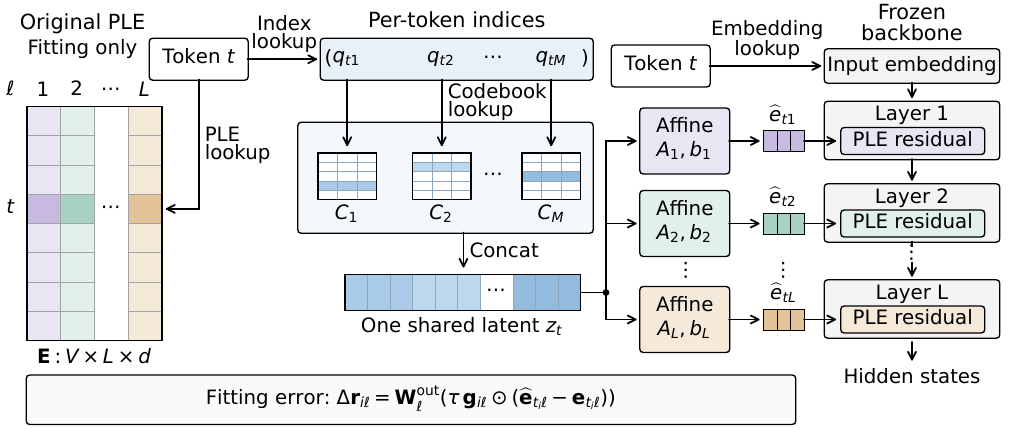}
\caption{One shared code per token serves all PLE layers.
Token $t$ retrieves its stored codebook indices and its input embedding.
Each index $q_{tm}$ selects the highlighted codeword of codebook $C_m$.
Concatenating these codewords forms one shared latent $\mathbf z_t$,
from which each affine decoder reconstructs its layer's PLE row.
Matching colors identify corresponding layers,
and ellipses mark omitted layers and codebooks. Each cell of $\mathbf E$
contains a $d$-dimensional vector. The original PLE is used only for fitting,
with frozen gates and projections (Equation~\ref{eq:ple-path}).
The frozen backbone shows the PLE residual connections and omits
attention and feed-forward operations.}
\label{fig:shared-codes}
\end{figure*}

Row-wise vector quantization stores a separate code tuple for each
token--layer pair $(t,\ell)$. OPTIC instead represents the
$L$ aligned PLE vectors of token $t$ with one code tuple shared across
layers, while all tokens of a layer share one affine decoder that models
the differences among layers (Figure~\ref{fig:shared-codes}).

Let $Z$ be the latent width and $M$ the number of equal-width codebook groups,
where $M$ divides $Z$ and $d_c=Z/M$ is the width of each group. For group
$m\in\{1,\ldots,M\}$, let
$\mathbf C_m\in\mathbb R^{K\times d_c}$ be a codebook containing $K$ codewords,
and let $q_{tm}\in\{0,\ldots,K-1\}$ select the codeword for token $t$. For layer
$\ell$, let $\mathbf A_\ell\in\mathbb R^{d\times Z}$ and
$\mathbf b_\ell\in\mathbb R^d$ be its affine decoder parameters. The compact
representation $\boldsymbol\theta$ contains these assignments, codebooks, and decoders,
and reconstructs a requested row as
\begin{equation}
\begin{aligned}
  \mathbf z_t &= \operatorname{concat}_{m=1}^{M}\mathbf C_m[q_{tm},:]
     \in\mathbb R^Z,\\
  \operatorname{Dec}_{\boldsymbol\theta}(t,\ell)
     &=\mathbf A_\ell\mathbf z_t+\mathbf b_\ell\in\mathbb R^d.
\end{aligned}
  \label{eq:latent-vq-decode}
\end{equation}
The latent width $Z$ describes the token representation shared across
layers; each reconstructed PLE row still has width $d$.
Let $b_q=\lceil\log_2K\rceil$ be the number of bits per assignment. The
assignment payload occupies
$\lceil VMb_q/8\rceil$ bytes, compared with
$\lceil VLMb_q/8\rceil$ bytes for a row-wise representation with the same
$M$ and $K$. The complete compact representation also stores the codebooks
and layer decoders.
With float16 codebooks and decoder parameters, the complete resident size is
\begin{equation}
  B_{\mathrm{latent}}
  =\left\lceil\frac{VMb_q}{8}\right\rceil+2KZ+2LdZ+2Ld.
  \label{eq:appendix-latent-bytes}
\end{equation}
The four terms account for token codes, codebooks, decoder weights and
decoder biases. Sharing codes removes the layer factor from the first
term. The decoder cost still grows with the number of layers, but it does
not grow with the vocabulary size because all tokens of a layer share its
decoder.

For E4B, $V=262{,}144$, $L=42$ and $d=256$. The configuration near
$100\times$ uses $Z=644$, $M=161$ and $K=256$, for a total of
56,404,992 bytes. The uncompressed bfloat16 PLE occupies
5,637,144,576 bytes, giving a compression ratio of $99.94\times$.
Decoding token $t$ at layer $\ell$ gathers only its $M$ codewords and applies
the corresponding layer decoder, without materializing the full
$V\times L\times d$ PLE tensor.
\subsection{Output-Path Fitting}
\label{sec:output-path-fitting}
The PLE affects the model only through a gated residual branch in each
decoder layer. This branch multiplies the PLE row element-wise by a gate
computed from the layer's hidden state and projects the product to the hidden
width. We fit the reconstruction at this branch output, before normalization
and residual addition.

Let $\mathcal I_{\mathrm{fit}}$ be a collection of $n$ fitting records. Record
$i\in\mathcal I_{\mathrm{fit}}$ contains token ID $t_i$ and a gate vector
$\mathbf g_{i\ell}\in\mathbb R^d$ for each layer $\ell$.
Natural-text records cache context-dependent gates from the uncompressed model,
whereas synthetic vocabulary records use all-ones gates
(Section~\ref{sec:fitting-data}). We hold each record's gates fixed during fitting.
Let
$\mathbf W^{\mathrm{out}}_\ell\in\mathbb R^{h\times d}$ be the frozen output
projection at layer $\ell$, where $h$ is the model hidden width, and let
$\tau>0$ be the fixed PLE scale. With the gate and all other inputs
fixed, replacing the original PLE row with its reconstruction changes the
layer's projection output by
\begin{equation}
  \Delta\mathbf r_{i\ell}(\boldsymbol\theta)
  =
  \mathbf W^{\mathrm{out}}_\ell
  \left(
    \tau\,\mathbf g_{i\ell}\odot
    \bigl(
      \widehat{\mathbf e}_{\boldsymbol\theta,t_i\ell}
      -\mathbf e_{t_i\ell}
    \bigr)
  \right)
  \in\mathbb R^h,
  \label{eq:ple-path}
\end{equation}
where $\odot$ denotes element-wise multiplication. We use the empirical
objective
\begin{equation}
  \mathcal L_{\mathrm{path}}(\boldsymbol\theta)
  =
  \frac{1}{nL}
  \sum_{i\in\mathcal I_{\mathrm{fit}}}
  \sum_{\ell=1}^{L}
  \left\|
    \Delta\mathbf r_{i\ell}(\boldsymbol\theta)
  \right\|_2^2.
  \label{eq:ple-path-loss}
\end{equation}
\subsection{Alternating Optimization}
Fitting alternates between continuous updates to codebooks and decoders
and discrete updates to token codes. With codes fixed, continuous updates
minimize Equation~\ref{eq:ple-path-loss}. Code reassignment updates the codes
of every token in the vocabulary, including tokens with few fitting
records. We therefore average the gate information over all fitting records
to obtain one error metric per layer:
\begin{equation}
\begin{aligned}
  \overline{\mathbf G}_\ell
    &=\bigl((\mathbf W^{\mathrm{out}}_\ell)^{\mathsf T}
              \mathbf W^{\mathrm{out}}_\ell\bigr)\\
    &\quad\odot\left(\frac{1}{n}\sum_{i\in\mathcal I_{\mathrm{fit}}}
                  \mathbf g_{i\ell}\mathbf g_{i\ell}^{\mathsf T}\right).
\end{aligned}
  \label{eq:latent-vq-marginal-metric}
\end{equation}
We drop the factor $\tau^2$ because it is positive and common to all
layers. For a positive-semidefinite matrix $\mathbf G$, define
$\|\mathbf x\|_{\mathbf G}^2=\mathbf x^{\mathsf T}\mathbf G\mathbf x$.
With the codebooks and decoders fixed, we use the following
surrogate for code reassignment:
\begin{equation}
\begin{aligned}
\widetilde{\mathcal L}_{\mathrm{assign}}(\boldsymbol\theta)
  &=\frac1V\sum_{t=1}^{V}\biggl[
    \sum_{\ell=1}^{L}
       \|\widehat{\mathbf e}_{\boldsymbol\theta,t\ell}-\mathbf e_{t\ell}
          \|_{\overline{\mathbf G}_\ell}^{2}\\
  &\qquad\qquad+\lambda\|\mathbf z_t\|_2^2\biggr].
\end{aligned}
  \label{eq:latent-vq-assignment}
\end{equation}
Here $\lambda\geq 0$ controls latent-norm regularization.
We update the codebook index of one group at a
time, choosing the codeword that minimizes
Equation~\ref{eq:latent-vq-assignment} while holding the other groups fixed.

Initialization uses the layer metrics $\overline{\mathbf G}_\ell$. We
concatenate the $L$ PLE rows of each token into one vector of width $Ld$,
subtract the mean of these vectors over the vocabulary, and scale the coordinates of layer $\ell$ by
$\overline{\mathbf G}_\ell^{1/2}$. These vectors, one per token, form a
$V\times Ld$ matrix. Principal component analysis (PCA) of
this matrix gives $Z$ leading directions. The projection of each
scaled vector onto these directions initializes the token latent
$\mathbf z_t$. The coordinates of layer $\ell$ in these directions, scaled by
$\overline{\mathbf G}_\ell^{-1/2}$, initialize the decoder $\mathbf A_\ell$,
and each layer's vocabulary mean initializes its bias $\mathbf b_\ell$. We assign the latent coordinates to
codebook groups by the eigenvalue allocation of optimized product
quantization (OPQ)~\citep{ge2013optimized}, so that the groups have similar
products of eigenvalues. We then run $K$-means within each group to
initialize its codewords and assignments.

Algorithm~\ref{alg:proposed-ple-methods} summarizes the procedure.
\hyperref[app:fitting-procedure]{Appendix~\ref*{app:fitting-procedure}}
gives the update schedule, unused-codeword handling and hyperparameters.
\begin{algorithm}[t]
  \caption{Fitting procedure for OPTIC.}
  \label{alg:proposed-ple-methods}

  \begin{algorithmic}[1]\small
    \RequireText{$\mathbf E$, $\mathcal I_{\mathrm{fit}}$, frozen gates and
      projections; $M$, $K$, and $Z$; $J$ continuous-update blocks of
      $U$ steps each}
    \Ensure Compact representation $\boldsymbol\theta$
    \StateText{Form the marginal layer metrics
      $\{\overline{\mathbf G}_\ell\}_{\ell=1}^{L}$
      (Equation~\ref{eq:latent-vq-marginal-metric})}
    \StateText{Initialize $\{\mathbf z_t\}$, $\{\mathbf A_\ell\}$ and
      $\{\mathbf b_\ell\}$ by PCA of the centered token vectors weighted by
      $\overline{\mathbf G}_\ell^{1/2}$}
    \StateText{Group the $Z$ coordinates by OPQ eigenvalue allocation}
    \StateText{Initialize $\{\mathbf C_m\}_{m=1}^{M}$ and
      $\{q_{tm}\}_{t,m}$ by $K$-means within each group}
    \For{$c=1,\ldots,J$}
      \For{$u=1,\ldots,U$}
        \StateText{Sample a minibatch of record--layer pairs and take one
          AdamW step on the codebooks and decoders to reduce
          $\mathcal L_{\mathrm{path}}$, with $\{q_{tm}\}$ fixed}
      \EndFor
      \If{$c<J$}
        \StateText{For every token, compute the unquantized minimizer
          $\mathbf z^\star_t$ of Equation~\ref{eq:latent-vq-assignment}}
        \StateText{For two sweeps over $m=1,\ldots,M$, set each $q_{tm}$ to
          the codeword of $\mathbf C_m$ that minimizes
          Equation~\ref{eq:latent-vq-assignment} with the other groups fixed}
        \StateText{Set each unused codeword to the group slice of $\mathbf z^\star_t$
          for a different worst-fit token, assign that token to the codeword and
          reset the codeword's AdamW moments}
      \EndIf
    \EndFor
    \State \Return $\boldsymbol\theta$
  \end{algorithmic}
\end{algorithm}
\subsection{Additional Compact Representations}
Output-path-weighted PCA-PQ stores one code tuple per token--layer row.
It scales each row with diagonal output-path weights and then applies PCA
and product quantization (PQ).
\hyperref[app:pca-pq]{Appendix~\ref*{app:pca-pq}} gives its derivation
and fitting procedure, and \hyperref[app:proposed-representations]{Appendix~\ref*{app:proposed-representations}}
gives its complete storage cost.

The frequency-tiered variant of OPTIC gives more code groups to frequent tokens.
Every token stores $M_0$ base groups, and the $N_{\mathrm{freq}}$ most frequent
non-special tokens in a fixed fitting stream store $M_1-M_0$ additional
groups. Other tokens use a learned shared codeword in each additional
group, so every vocabulary item keeps an explicit code. The variant also
stores INT8 decoder weights and keeps special-token rows exactly in
bfloat16. \hyperref[app:proposed-representations]{Appendix~\ref*{app:proposed-representations}}
gives its complete storage cost.
\section{Experimental Setup}
\label{sec:experiments}
\subsection{Baselines}
We evaluate six baselines adapted to the PLE setting: flattened row-wise
product quantization (PQ)~\citep{jegou2010product}, greedy residual vector
quantization (RVQ)~\citep{chen2010approximate}, parametric optimized product
quantization (OPQ-P)~\citep{ge2013optimized}, Additive Quantization
(AQ)~\citep{babenko2014additive}, the AQLM Euclidean beam
encoder~\citep{egiazarian2024extreme}, and post-hoc INT8 randomized low-rank
singular value decomposition (SVD)~\citep{halko2011finding}. Our methods are
output-path-weighted PCA-PQ and OPTIC.
Each baseline is applied to the PLE tensor under matched resident-byte accounting,
and \hyperref[app:baselines]{Appendix~\ref*{app:baselines}} gives its
representation and fitting rule.

\subsection{Calibration Data}
\label{sec:fitting-data}
The calibration data for OPTIC and its frequency-tiered variant combine natural
text with synthetic records. The natural text
consists of 1,536 sequences of 512 tokens (786,432 positions) from FineWeb and
FineWeb-Edu~\citep{penedo2024fineweb}, OpenWebMath~\citep{paster2024openwebmath}
and FineWeb2~\citep{penedo2025fineweb2}. Because the natural text covers only
part of the vocabulary, we add one synthetic record for every token in the
vocabulary. Repeated special-token records give the fitting procedure
additional coverage of tokens used to control generation. Natural records
use gates collected from the frozen model, whereas synthetic records use
all-ones gates with the same frozen projections.

The frequency-tiered variant uses token counts from a fitting stream that
combines these text sources with text generated by the uncompressed
checkpoint. Output-path-weighted PCA-PQ uses the WikiText-2 training
split~\citep{merity2016pointer} for its fitting statistics. Calibration data
exclude evaluation examples.

\begin{table*}[t]
  \centering
  \resizebox{0.75\textwidth}{!}{%
  \begin{tabular}{@{}lcccc@{}}
    \toprule
    & \multicolumn{2}{c}{Original BF16} &
      \multicolumn{2}{c}{Official W4A16} \\
    \cmidrule(lr){2-3}\cmidrule(lr){4-5}
    Method & \shortstack{E2B\\$R$ / Retention} & \shortstack{E4B\\$R$ / Retention}
           & \shortstack{E2B\\$R$ / Retention} & \shortstack{E4B\\$R$ / Retention} \\
    \midrule
    \multicolumn{5}{l}{\emph{Baselines}} \\
    PQ      & 101.24/22.9 & 101.43/66.3 & 101.24/0.0  & 101.43/82.4 \\
    RVQ     & 100.96/0.0  & 98.64/37.6  & 100.96/29.3 & 101.20/82.0 \\
    OPQ-P   & 100.92/0.0  & 101.16/52.4 & 100.92/0.0  & 101.16/57.9 \\
    AQ      & 100.96/0.0  & 101.20/45.6  & 100.96/0.0  & 101.20/62.8 \\
    AQLM Euclidean beam encoder
      & 100.96/20.7 & 101.20/46.7 & 100.96/32.0 & 101.20/82.8 \\
    INT8 randomized low-rank SVD
      & 100.13/0.0  & 99.76/37.7  & 100.13/0.0  & 100.24/51.1 \\
    \midrule
    \multicolumn{5}{l}{\emph{Our methods}} \\
    \Owpq{}
      & 96.14/32.2 & 96.36/93.3 & 96.14/44.5 & 96.36/88.3 \\
    OPTIC
      & 99.86/89.1 & 99.94/\textbf{97.8}
      & 99.86/91.5 & 99.94/\textbf{96.9} \\
    Frequency-tiered OPTIC
      & 100.48/\textbf{90.0} & 100.00/95.7
      & 100.48/\textbf{93.0} & 100.00/95.1 \\
    \bottomrule
  \end{tabular}%
  }
  \caption{PLE compression on the original BF16 and official W4A16 releases
  at the nominal $100\times$ tier. Cells report compression ratio
  $R$ / seven-task retention (\%) relative to the matched uncompressed PLE.
  A zero denotes at least one zero task score. Bold marks the highest retention
  in each column.}
  \label{tab:primary-matrices}
\end{table*}
\subsection{Evaluation Protocol}
The seven tasks cover reasoning with PIQA~\citep{bisk2020piqa},
GSM8K~\citep{cobbe2021training}, GPQA
Diamond~\citep{rein2023gpqa} and AIME 2026~\citep{dekoninck2026beyond};
world knowledge with MMLU-Redux~\citep{hendrycks2020measuring,gema2025we};
instruction following with IFBench~\citep{pyatkin2026generalizing}; and
vision with InfographicVQA~\citep{mathew2022infographicvqa}.
For each model size and weight format, all methods use the same backbone
weights, prompts, generation settings and scorers. The main comparisons use fitting seed 0, and every fit is evaluated after
its last scheduled update.
\hyperref[app:models-evaluation]{Appendix~\ref*{app:models-evaluation}}
gives sample counts and evaluation seeds.

For method $m$ and task $k$, let $s_{m,k}$ be its score and
$s_{\mathrm{ref},k}$ the score of the same checkpoint with uncompressed PLE.
We report the equal-weight geometric mean of capped per-task retention:
\begin{equation}
  100\left[\prod_{k=1}^{7}
  \min\left(\frac{s_{m,k}}{s_{\mathrm{ref},k}},1\right)\right]^{1/7}.
  \label{eq:retention}
\end{equation}
Capping prevents gains above the reference on one task from offsetting
losses on another. The geometric mean becomes zero if any task score is zero.

For W4A16 experiments, each compact PLE artifact is fitted from the matched
unquantized checkpoint from the official quantization-aware-training pipeline
and applied to its W4A16
release~\citep{google2026gemma4e2bw4a16,google2026gemma4e4bw4a16}, whose PLE lookup
remains in bfloat16.
\section{Compression Results}
\subsection{Compression and Quality}
Table~\ref{tab:primary-matrices} compares methods near $100\times$ PLE
compression. OPTIC gives the highest E4B retention:
97.8\% on BF16 and
96.9\% on W4A16. These E4B results lead the strongest baseline by 31.5 and
14.0 percentage points, respectively. The frequency-tiered variant of OPTIC
gives the highest E2B retention: 90.0\% on BF16 and 93.0\% on W4A16.

The per-task scores show where the E4B gains arise. On BF16 E4B,
OPTIC preserves GSM8K accuracy at 92.8\%, compared with
92.9\% for uncompressed PLE and 73.5\% for PQ. Its IFBench score is
41.7\%, compared with 44.0\% for uncompressed PLE and 16.0\% for PQ.

InfographicVQA follows the same pattern: 67.5\% for OPTIC,
69.7\% for uncompressed PLE and 38.4\% for PQ. PIQA changes little for
either compressed method. The aggregate advantage includes gains
in mathematical reasoning, instruction following and visual question
answering. Appendix~\ref{app:results} reports every task score.

Table~\ref{tab:appendix-headline-bootstrap} gives 95\% paired-bootstrap
intervals. The E4B BF16 interval is
95.5--99.0\% for OPTIC and 60.6--70.5\% for PQ.
The corresponding W4A16 intervals are 93.9--97.8\% and 76.8--87.1\%
for OPTIC and the AQLM Euclidean beam encoder.
The intervals remain separated in both E2B settings as well.
Appendix~\ref{app:results} specifies resampling.

\begin{table*}[t]
  \centering
  \resizebox{0.79\textwidth}{!}{%
  \begin{tabular}{@{}llclc@{}}
    \toprule
    Setting & Our method & \shortstack{Retention\\(95\% interval)}
      & Strongest baseline & \shortstack{Retention\\(95\% interval)} \\
    \midrule
    E2B, BF16
      & OPTIC
      & 89.1 [85.8, 91.8] & PQ & 22.9 [0.0, 28.8] \\
    E4B, BF16
      & OPTIC
      & 97.8 [95.5, 99.0] & PQ & 66.3 [60.6, 70.5] \\
    E2B, W4A16
      & OPTIC
      & 91.5 [89.3, 93.3] & AQLM Euclidean beam encoder
      & 32.0 [0.0, 38.9] \\
    E4B, W4A16
      & OPTIC
      & 96.9 [93.9, 97.8] & AQLM Euclidean beam encoder
      & 82.8 [76.8, 87.1] \\
    \bottomrule
  \end{tabular}
  }
  \caption{Paired-bootstrap confidence intervals for the four primary
  nominal-$100\times$ comparisons. Retention is reported in percent.}
  \label{tab:appendix-headline-bootstrap}
\end{table*}

\begin{figure*}[t]
\centering
\includegraphics[width=\textwidth]{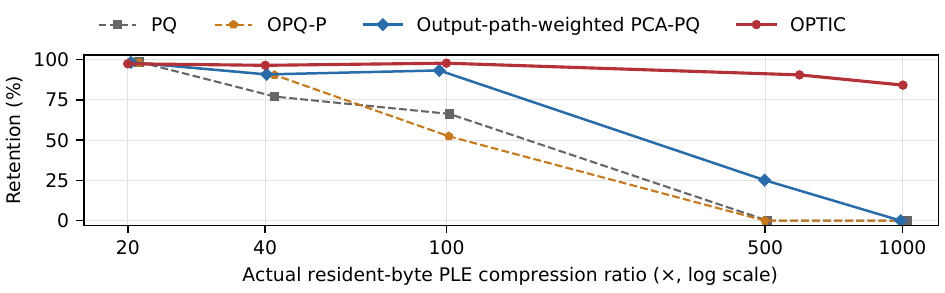}
\caption{Rate--retention curves on BF16 Gemma~4 E4B, normalized to the
  uncompressed PLE. Dashed curves show the PQ and OPQ-P baselines, and solid curves
  show our methods. OPTIC retains 97.8\% near
  $100\times$, and its advantage over the shown baselines remains large at
  higher compression ratios. A zero geometric mean means that at least one
  task scored zero.}
\label{fig:bf16-rate-curve}
\end{figure*}

\begin{table*}[!t]
\centering
\resizebox{0.63\textwidth}{!}{%
\begin{tabular}{lrrrr}
\toprule
Representation & E2B BF16 & E4B BF16 & E2B W4A16 & E4B W4A16 \\
\midrule
Independent latent codes & 28.5 & 79.3 & 53.7 & 86.8 \\
Additive codes & 42.1 & 90.0 & 83.5 & 94.4 \\
OPTIC & 89.1 & 97.8 & 91.5 & 96.9 \\
\bottomrule
\end{tabular}%
}
\caption{Seven-task retention (\%) near $100\times$ PLE compression with fit
seed 0, relative to the uncompressed checkpoints of
Tables~\ref{tab:appendix-bf16-per-task} and~\ref{tab:appendix-w4a16-per-task}.
Independent latent codes use the main method's fitting recipe and calibration
data, with 0.8--0.9\% fewer resident bytes. OPTIC values are
the main results of Table~\ref{tab:primary-matrices}. Output-path-fitted additive codes keep
their standard initializer.}
\label{tab:sharing-ablation}
\end{table*}
\subsection{Compression Range}
Figure~\ref{fig:bf16-rate-curve} compares OPTIC with PQ,
OPQ-P and output-path-weighted PCA-PQ across five nominal tiers on BF16 E4B.
OPTIC has compression ratios of $594\times$ and $1002\times$ at the nominal
$500\times$ and $1000\times$ tiers. At these ratios, it retains 90.5\% and
84.1\% on BF16 E4B and 87.7\% and 81.8\% on W4A16 E4B
(Table~\ref{tab:appendix-e4b-extreme-rate}).

Output-path-weighted PCA-PQ is an output-aware row-wise method. It retains
93.3\% near $100\times$ on BF16 E4B, 25.1\% near $500\times$ and zero near
$1000\times$, while OPTIC remains above 84\%. The advantage of OPTIC over
this row-wise method therefore grows as fewer bytes are available
to represent each PLE vector.

\section{Analysis and Ablations}
\label{sec:ablations}
This section examines the structure of the PLE table
(Section~\ref{sec:ple-structure}), the effect of sharing token codes across
layers (Section~\ref{sec:code-sharing}) and the components of the fitting
procedure (Section~\ref{sec:fitting-components}).
\subsection{PLE Structure}
\label{sec:ple-structure}
We examine the importance and dimensional structure of token-specific
variation in PLE, using the original BF16 E4B table with equal weighting
over vocabulary items (Appendix~\ref{app:cross-layer-structure}). Each
vector $\mathbf e_{t\ell}$ splits into its layer's vocabulary mean and a
centered remainder. The means account for 8.1\% of the table's squared
Frobenius norm. Replacing every vector with its layer mean lowers GSM8K
from 92.9\% to 43.5\% and GPQA Diamond from 56.9\% to 23.0\%; AIME~2026
falls to zero, making seven-task retention zero. Layer means alone therefore do not preserve model performance.

\begin{figure}[t]
\centering
\includegraphics[width=\linewidth]{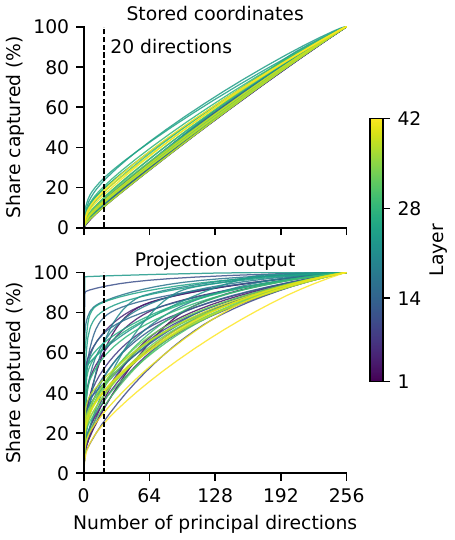}
\caption{Cumulative share of each layer's token-specific variation
captured by its leading principal directions in the original BF16 E4B PLE
table, with every token in the vocabulary weighted equally, in the stored
coordinates (top) and at the PLE projection output (bottom). Each curve is
one of the 42 layers. The dashed line marks 20 directions, the latent width
of independent latent codes (Section~\ref{sec:code-sharing}).}
\label{fig:ple-spectra}
\end{figure}

We next examine whether the remaining variation is concentrated in a
low-dimensional subspace. Besides the stored coordinates, we use the
output-path metric $\overline{\mathbf G}_\ell$
(Equation~\ref{eq:latent-vq-marginal-metric}) to account for the gate and
projection before normalization. In stored coordinates, capturing 90\%
of a layer's token-specific variation requires 212--227 of 256 principal
directions. Under the output-path metric, it requires 1--205 directions
(median 158), indicating substantial differences across layers
(Figure~\ref{fig:ple-spectra}).

\subsection{Code Sharing}
\label{sec:code-sharing}
We examine how cross-layer code sharing affects performance at nearly
equal storage. Independent latent
codes assign a separate code tuple to each token--layer pair and use
layer-specific affine decoders. We fit them with the main method's recipe,
including output-path-weighted initialization, unused-codeword handling,
calibration data and fit seed. Appendix~\ref{app:controlled-fitting} gives
the configurations.

On E4B, OPTIC retains 97.8\% with BF16 and 96.9\% with W4A16,
compared with 79.3\% and 86.8\% for independent latent codes
(Table~\ref{tab:sharing-ablation}). Three-fit means under a common
Euclidean initialization, which applies PCA and $K$-means without
output-path weights, show the same ordering
(Table~\ref{tab:main-comparison}).

Near $100\times$, OPTIC stores 161 one-byte codebook indices per token,
shared across layers, while independent latent codes store five per
token--layer pair. Each index selects a four-dimensional codeword, giving
latent widths of $161\times4=644$ and $5\times4=20$, respectively.
To test the role of decoder rank, a decoder-rank control limits every
layer decoder to rank 20 (Table~\ref{tab:sharing-controls}). It keeps the
indices, codebooks and biases of the three common-initialization E4B BF16
fits of Table~\ref{tab:main-comparison} and refits only decoder weights by
frequency-weighted least squares on the PLE rows. With this objective,
refits capped at rank 20 retain 70.55\%, compared with 95.09\% without the limit.
Reducing decoder rank lowers retention even when the shared codes
are~unchanged.

As an additional comparison without code sharing, output-path-fitted
additive codes reconstruct
each row as a sum of codewords and keep their standard initializer. They
retain 90.0\% on BF16 E4B and 94.4\% on W4A16 E4B, below OPTIC.

\begin{table}[!t]
\centering
\resizebox{0.84\linewidth}{!}{%
\begin{tabular}{lr}
\toprule
Decoder refit with fixed codes & Retention (\%) \\
\midrule
Rank unlimited & $95.09\pm0.48$ \\
Fitted rank at most 20 per layer & $70.55\pm1.18$ \\
\bottomrule
\end{tabular}%
}
\caption{Decoder refits on the three E4B BF16 fits of
Table~\ref{tab:main-comparison}, which use a common Euclidean
initialization, near $100\times$. Values are means and sample standard
deviations over fit seeds 0, 1 and 2. Refits keep each fit's indices,
codebooks and biases, constrain rank before float16 storage, and share the
uncompressed reference of Table~\ref{tab:main-comparison}.}
\label{tab:sharing-controls}
\end{table}

\subsection{Fitting Components}
\label{sec:fitting-components}
\begin{table}[!ht]
\centering
\resizebox{0.83\linewidth}{!}{%
\begin{tabular}{lr}
\toprule
Fixed-code objective & Retention (\%) \\
\midrule
Euclidean & $92.76\pm1.16$ \\
Diagonal output path & $94.44\pm0.94$ \\
Context-dependent output path & $94.74\pm0.59$ \\
\bottomrule
\end{tabular}%
}
\caption{Fixed-code fitting on E4B BF16 near $100\times$, with the common Euclidean initialization and 5,000 continuous updates. Each loss is divided by its value for an all-zero reconstruction. Values are means and sample standard deviations over three fits.}
\label{tab:objective}
\end{table}
We first isolate the fitting objective by fixing codes and initialization
and using 5,000 continuous updates. We divide each loss by its
value for an all-zero reconstruction and use the same learning rate and
gradient-clipping threshold for all three objectives. The context-dependent
objective gains 1.98 percentage points in paired mean retention over
Euclidean fitting and 0.31 points over diagonal weighting
(Table~\ref{tab:objective}). Output-path fitting therefore improves over
Euclidean fitting, and diagonal weighting captures most of this improvement.

\begin{table}[!ht]
\centering
\resizebox{0.91\linewidth}{!}{%
\begin{tabular}{lr}
\toprule
Optimization recipe & Retention (\%) \\
\midrule
Initialization only & $63.60\pm3.68$ \\
Continuous updates only & $94.67\pm0.53$ \\
Continuous updates + reassignment & $95.49\pm0.40$ \\
\bottomrule
\end{tabular}%
}
\caption{Optimization on E4B BF16 near $100\times$. Values are means and sample standard deviations over three fits. The last two rows use 5,000 continuous updates, and the final row adds 18 reassignment sweeps.}
\label{tab:optimization}
\end{table}
We then assess continuous fitting and code reassignment from the common
Euclidean initialization (Table~\ref{tab:optimization}). Continuous
fitting supplies most of the gain; reassignment improves all three paired
fits, with paired gains of 0.65--1.14 percentage points
(Appendix~\ref{app:optimization}).

\section{Compact Inference}
\label{sec:compact-inference}
The compression ratios above measure PLE storage. We now measure the
memory and runtime of the full model when serving directly from the
compact representation.
We serve uncompressed (dense) and approximately $100\times$ compact PLE for BF16 and
W4A16 E4B with vLLM~\citep{kwon2023efficient} on one
RTX~4090. Compact PLE keeps only the codes, codebooks and layer decoders on
the GPU and decodes the rows of the current tokens at each step. Each pair
uses identical backbone weights and serving settings, including a fixed
1~GiB key--value cache (Table~\ref{tab:runtime-memory}).

\begin{table}[!ht]
\centering
\resizebox{0.95\linewidth}{!}{%
\begin{tabular}{lrrr}
\toprule
Backbone & Dense (GB) & Compact (GB) & Saved (\%) \\
\midrule
BF16 & 18.57 & 13.05 & 29.7 \\
W4A16 & 12.95 & 7.37 & 43.1 \\
\bottomrule
\end{tabular}%
}
\caption{Post-initialization process GPU memory for E4B on an RTX~4090 with a fixed 1~GiB key-value cache, averaged over two engine initializations. Resident PLE storage falls from 5.64~GB to 56.4~MB.}
\label{tab:runtime-memory}
\end{table}
Compact lookup adds at most 0.40\% wall time with BF16 and 0.69\% with W4A16
across eight prefill and generation workloads.
Appendix~\ref{app:runtime} also reports serving at a fixed GPU memory
budget, single long requests and PLE lookups with the table stored
in host memory.

\section{Related Work}

\paragraph{Embedding compression.}
Multi-codebook representations have been used to compress word embeddings
for NLP. \citet{shu2018compressing} learn compositional discrete codes and
codebooks for an embedding matrix. Our setting has an additional layer axis:
the same vocabulary item has a different embedding at each layer. We exploit
this correspondence by decoding one token representation into multiple
layer-specific rows. CARVQ combines grouped residual codes with a nonlinear
corrective adaptor for ordinary token embeddings~\citep{gou-etal-2025-carvq};
our representation instead shares a token's latent codes across layers.

\paragraph{Vector quantization and low-rank representations.}
Product quantization divides a vector into subspaces and quantizes each
subspace~\citep{jegou2010product}; optimized product quantization learns a
transformation before quantization~\citep{ge2013optimized}.
Residual and additive quantizers represent vectors through sums of
codewords~\citep{chen2010approximate,babenko2014additive}.
Low-rank factorization instead stores continuous coordinates and a shared
basis; randomized SVD offers a practical fitting
procedure~\citep{halko2011finding}. OPTIC combines a shared
latent representation with discrete token codes and layer-specific
affine decoders. This shares the code storage across layers while keeping a separate
reconstruction map for each layer.

\paragraph{Output-aware model compression.}
Weight reconstruction can account for a model's inputs and activations.
GPTQ uses approximate second-order information for post-training weight
quantization~\citep{frantar2022gptq}, while AWQ uses activation statistics
to choose channel scales that protect salient
weights~\citep{lin2024awq}.
AQLM combines input-adaptive additive quantization with joint block
optimization~\citep{egiazarian2024extreme}. Our loss follows the gated PLE
output path instead of a linear weight matrix. Compressing this separate
embedding table complements backbone weight quantization, as our BF16
and W4A16 experiments demonstrate.

Our comparisons include the AQLM Euclidean beam encoder and
output-path-fitted additive codes; Appendices~\ref{app:methods} and~\ref{app:controlled-fitting}
give their PLE adaptations and storage budgets.

\section{Conclusion}
OPTIC stores one code tuple per token, which every layer decodes through its
own affine decoder, and fits codes and decoders at the PLE projection
output, before normalization. Near $100\times$ PLE compression, it achieves
seven-task retention of 97.8\% on BF16 E4B and 96.9\% on W4A16 E4B. Independent latent codes
for each vector retain 79.3\% on BF16 E4B with nearly equal storage.
At $594\times$ and $1002\times$ PLE compression, OPTIC retains
90.5\% and 84.1\% of BF16 E4B performance, respectively.
Served directly from OPTIC codes with a fixed key--value cache size,
BF16 E4B uses 29.7\% less GPU memory with at most 0.40\% added wall time.

These results show substantial compression potential in Gemma~4's expanded
embedding tables. They motivate designing embedding capacity and compressed
representations together, with deployment memory as a constraint.
Our BF16 and W4A16 results show that PLE compression complements backbone
quantization; extending this combination to other embedding structures
is a promising direction.

\clearpage
\section*{Limitations}
The study evaluates PLE in Gemma~4 E2B and E4B.
The main comparisons and Table~\ref{tab:sharing-ablation} use one fixed
fitting seed, so the paired-bootstrap intervals describe benchmark sampling
uncertainty. The three-fit comparisons and controls describe fitting variation under
the common Euclidean initialization.

At a matched byte budget, removing code sharing also changes how bytes are
divided between codes and decoders, so the code-sharing ablation measures this
combined effect with the fitting recipe held fixed. The fixed-code objective ablation applies
to continuous fitting, and the output-path loss holds the gates fixed and
measures error only at each layer's projection output. The baselines, including the
AQLM Euclidean beam encoder, are adaptations of their original methods to PLE. A single low-count task can strongly affect retention.

\section*{Ethical Considerations}
Compression reduces the resources needed to study and deploy existing
models. It can broaden access to model research, although the same
resource savings can also lower the cost of harmful applications.
AI assistants were used for language polishing and code assistance.
The authors remain responsible for the results, references,
implementation and writing.

\section*{Acknowledgments}

This work is supported by the MBZUAI-WIS Joint Program for Artificial Intelligence Research.

\bibliography{references}

\clearpage
\appendix
\section*{Appendix}
\noindent\hyperref[app:methods]{Method Details} specifies representations and storage.
\hyperref[app:protocol]{Implementation and Protocol} gives fitting, checkpoints and serving.
\hyperref[app:results]{Additional Results} reports raw scores, rate comparisons and uncertainty.
\hyperref[app:pca-pq]{Output-Path-Weighted PCA-PQ} gives its fitting procedure and derivation.
\hyperref[app:controlled-fitting]{Controlled Comparisons} specifies the supplementary experiments.
\section{Method Details}
\label{app:methods}

\subsection{Notation and Storage}

For model size $a\in\{\mathrm{E2B},\mathrm{E4B}\}$, let $V=262{,}144$ be
the vocabulary size, $d=256$ the PLE-vector dimension, and $L_a$ the number
of decoder layers, with $L_{\mathrm{E2B}}=35$ and
$L_{\mathrm{E4B}}=42$. The PLE tensor therefore contains $N_a=VL_a$
token--layer rows. Its uncompressed size $B_{\mathrm{ref},a}$ and the
reported compression ratio $R_a$ are
\begin{equation}
  B_{\mathrm{ref},a}=2VL_a d,
  \qquad
  R_a=\frac{B_{\mathrm{ref},a}}{B_{\mathrm{method},a}},
  \label{eq:appendix-compression-ratio}
\end{equation}
where $B_{\mathrm{method},a}$ includes packed codes and every persistent
codebook, transform, scale, exception, and decoder tensor. The official W4A16
releases keep their PLE values in bfloat16 and thus use the same PLE reference
size as the corresponding original BF16 model. Both benchmark evaluation
and compact serving use the reconstruction in Equation~\ref{eq:latent-vq-decode}.
Benchmark evaluation materializes the reconstructed PLE; the serving
experiment in Section~\ref{app:runtime} computes only requested rows from
the resident compact representation. We omit the size subscript below when a
definition applies to both models.

\subsection{Baselines}
\label{app:baselines}

We adapt six established methods to the PLE setting by operating on the
PLE tensor and matching the complete resident-state budget.
PQ~\citep{jegou2010product} partitions each row into independent subspaces;
RVQ~\citep{chen2010approximate} greedily quantizes the full-vector
residual; and OPQ-P~\citep{ge2013optimized,he2013opqcode} applies its learned
rotation before PQ. AQ~\citep{babenko2014additive,babenko2015quantizations}
uses the authors' beam assignment and alternating least-squares codebook
update. The AQLM Euclidean beam encoder
~\citep{egiazarian2024extreme,aqlmauthors2026code} uses its source Euclidean
encoder with $K=256$, beam width two, and a codebook count set by the byte
budget; the nominal-$100\times$ configuration uses five codebooks. Finally,
INT8 randomized low-rank SVD~\citep{halko2011finding} factorizes the
$V\times Ld$ matrix formed
by concatenating each token's layer vectors and stores both factors in INT8
with float16 scales. Packed assignments, codebooks, rotations, factor scales,
and other persistent tensors are included in every reported byte count.

\subsection{Proposed Representations}
\label{app:proposed-representations}

\paragraph{Output-path-weighted PCA-PQ.}
Section~\ref{app:pca-pq} defines the output-path weighting, rotation, encoder,
and decoder. The
residual set is selected from calibration-token frequency; its corrections
are stored in bfloat16, after which a layer-coordinate scale is fitted. Let
$M$ be the number of subspaces, $K$ the codewords per codebook,
$b=\lceil\log_2K\rceil$, $P$ the number of shared codebook banks, and $S$ the
number of residual-corrected tokens. The reported configuration uses one
shared bank, $P=1$. The complete compact representation occupies
\begin{equation}
\begin{aligned}
B_{\mathrm{path\text{-}PQ}}
  &=\left\lceil\frac{NMb}{8}\right\rceil+2PKd+2d^2\\
  &\quad+2d+4Ld+4S+2SLd.
\end{aligned}
  \label{eq:appendix-owpq-bytes}
\end{equation}

\paragraph{OPTIC.}
Section~\ref{sec:methods} defines the shared token codes, layer-specific affine decoders,
output-path loss, and assignment objective. Let $Z$ be the latent width, $M$
the number of code groups, $K$ the codewords per group, and
$b_q=\lceil\log_2K\rceil$. Equation~\ref{eq:appendix-latent-bytes} gives the complete resident size.
The assignment term scales with $V$, rather than $VL$, because every layer
shares a token's code tuple.

\paragraph{Frequency-tiered OPTIC.}
The variant orders latent coordinates by principal-component importance and
stores $M_0$ base groups for every token and $M_1-M_0$ additional groups for
the $N_{\mathrm{freq}}$ most frequent non-special tokens in a fixed fit-only
stream. Other tokens use a learned shared codeword (codeword zero) in each
additional group. With group width $d_c$, latent width $Z=M_1d_c$, and
$N_{\mathrm{sp}}$ special-token IDs whose $L$ PLE rows are retained exactly
in bfloat16, the artifact contains one-byte indices ($K=256$), float16
codebooks, INT8 decoder weights with float16 scales and biases, and 32-bit
frequent-token IDs. Its
complete resident size is
\begin{equation}
\begin{aligned}
B_{\mathrm{tiered}}
 &=VM_0+N_{\mathrm{freq}}(M_1-M_0)+4N_{\mathrm{freq}}\\
 &\quad+2M_1Kd_c+LdZ+4Ld+2N_{\mathrm{sp}}Ld.
\end{aligned}
  \label{eq:appendix-tiered-bytes}
\end{equation}
Near $100\times$, E2B uses $M_0=136$ and $M_1=170$, E4B uses $M_0=160$ and
$M_1=200$, and both use $d_c=4$, $N_{\mathrm{freq}}=110{,}411$ and
$N_{\mathrm{sp}}=24$.

\section{Implementation and Protocol}
\label{app:protocol}
The released code provides the environment, fitting implementations,
and evaluation code.

\subsection{Fitting Procedure}
\label{app:fitting-procedure}

OPTIC and its frequency-tiered variant use seed 0,
PCA initialization weighted by the output-path objective, 25 iterations of
$K$-means codebook initialization, and a fixed 5,000-step AdamW schedule.
The BF16 E4B nominal-$100\times$ configuration has latent width 644,
161 groups, 256 codewords per group and codeword width four.
Continuous updates use learning rate $3\cdot10^{-4}$, zero weight decay,
batches of 8,192 positions and gradient clipping at norm one.
We set $\lambda$ in Equation~\ref{eq:latent-vq-assignment} to $10^{-8}$
times the mean diagonal entry of
$\sum_{\ell}\mathbf A_\ell^{\mathsf T}\overline{\mathbf G}_\ell\mathbf A_\ell$.
Two reassignment sweeps follow updates 500, 1000, \ldots, 4500.
After reassignment, unused codewords are initialized from tokens
with large quantization errors, and their optimizer moments are reset.
Each fit runs all 5,000 updates, and the state after the last update is
the reported artifact.

\subsection{Models and Evaluation}
\label{app:models-evaluation}

The experiment crosses two Gemma 4 scales~\citep{team2026gemma} with the
original BF16 releases~\citep{google2026gemma4e2bit,google2026gemma4e4bit}
and official W4A16
releases~\citep{google2026gemma4e2bw4a16,google2026gemma4e4bw4a16}.
The W4A16 backbone quantizes linear weights while its PLE lookup remains in
bfloat16. For W4A16, each compact PLE artifact is fitted using the corresponding
unquantized checkpoint from the official quantization-aware-training
pipeline~\citep{google2026gemma4e2bqatparent,google2026gemma4e4bqatparent} and then
loaded into the official W4A16 release.

\begin{table*}[t]
  \centering
  \resizebox{0.8\textwidth}{!}{%
  \begin{tabular}{@{}llll@{}}
    \toprule
    Scale & Role & Hugging Face model & Version \\
    \midrule
    E2B & Evaluated BF16 & \path{google/gemma-4-E2B-it} & \texttt{b446025c61ec} \\
    E4B & Evaluated BF16 & \path{google/gemma-4-E4B-it} & \texttt{ee0ef6023621} \\
    E2B & PLE fit for W4A16 & \path{google/gemma-4-E2B-it-qat-q4_0-unquantized} & \texttt{6befbaca7398} \\
    E4B & PLE fit for W4A16 & \path{google/gemma-4-E4B-it-qat-q4_0-unquantized} & \texttt{476025a01dbf} \\
    E2B & Evaluated W4A16 & \path{google/gemma-4-E2B-it-qat-w4a16-ct} & \texttt{971342c08f60} \\
    E4B & Evaluated W4A16 & \path{google/gemma-4-E4B-it-qat-w4a16-ct} & \texttt{6cd26aaa2357} \\
    \bottomrule
  \end{tabular}
  }
  \caption{Model releases used for fitting and evaluation. The final column
  gives a 12-character revision prefix.}
  \label{tab:appendix-model-releases}
\end{table*}

We evaluate PIQA validation accuracy~\citep{bisk2020piqa} (1,838 examples),
eight-shot GSM8K test accuracy~\citep{cobbe2021training} (1,319 examples), GPQA
Diamond~\citep{rein2023gpqa} (198 questions under four fixed option
permutations; 792 attempts), AIME 2026~\citep{dekoninck2026beyond} (30
problems, four generations each; 120 attempts), IFBench~\citep{pyatkin2026generalizing} (300 prompts), InfographicVQA
validation~\citep{mathew2022infographicvqa} (2,801 questions), and
MMLU accuracy on the 3,000-question MMLU-Redux v1 subset, 30 subjects with 100
questions each~\citep{hendrycks2020measuring,gema2025we}.

For each model size and weight format, all methods use the same model weights,
prompts, generation settings, and scorers. The main comparisons use fitting seed 0. PIQA, GSM8K,
GPQA Diamond, and MMLU-Redux use evaluation seed 0; IFBench and
InfographicVQA use seed 42; AIME micro-averages all 120 attempts from seeds
0--3.

Every repeated attempt remains in its benchmark denominator. For method $m$
and benchmark $k$, let $s_{m,k}$ be the compressed-model score and
$s_{\mathrm{ref},k}$ the score of the same checkpoint and weight format with uncompressed
PLE. Retention is
\begin{equation}
  100\left[\prod_{k=1}^{7}
  \min\left(\frac{s_{m,k}}{s_{\mathrm{ref},k}},1\right)\right]^{1/7},
  \label{eq:appendix-seven-task}
\end{equation}
with the reference matched by model size and weight format.

\subsection{Compact Serving}
\label{app:runtime}

We benchmark the nominal-$100\times$ default OPTIC with BF16
and official W4A16 E4B separately on one RTX~4090 using
vLLM 0.19.1~\citep{kwon2023efficient,vllm2026v0191}. Each matched pair uses
identical non-PLE tensors and serving settings, including a fixed 1~GiB
key--value cache. The compact PLE representation keeps only codes, codebooks, and layer
decoders resident and decodes requested rows without materializing the full
PLE tensor. The stacked decoder uses float16 inputs with float32
accumulation and bias addition, then casts the reconstructed rows to
bfloat16. Both conditions use compiled execution with prefix caching
disabled and identical synthetic token prompts. We report
both the post-initialization process memory recorded after engine setup and the
logical bytes of resident PLE state; the former includes model weights,
the cache, and runtime allocations. Two
counterbalanced engine initializations provide six timed measurements per
workload after warm-up.
Prefill covers input lengths from 128 to 8,192 tokens; generation covers
batch sizes from one to 32 and output lengths from 32 to 512 tokens.
These workloads exercise compact lookup during both prompt processing
and repeated decoding.
\begin{table*}[!t]
  \centering
\resizebox{0.8\textwidth}{!}{%
  \begin{tabular}{@{}llrrrcc@{}}
    \toprule
    Weights & Phase & Batch & Input tokens & Output tokens & Wall time $\Delta$ (\%)
          & Throughput $\Delta$ (\%) \\
    \midrule
    BF16 & Prefill & 16 & 128   & 1   & -0.23 & +0.19 \\
    BF16 & Prefill & 4  & 512   & 1   & -0.55 & +0.55 \\
    BF16 & Prefill & 1  & 2,048 & 1   & +0.16 & -0.16 \\
    BF16 & Prefill & 1  & 8,192 & 1   & +0.31 & -0.31 \\
    BF16 & Decode  & 32 & 128   & 32  & +0.17 & -0.32 \\
    BF16 & Decode  & 16 & 128   & 128 & +0.20 & -0.30 \\
    BF16 & Decode  & 4  & 128   & 512 & +0.32 & -0.35 \\
    BF16 & Decode  & 1  & 2,048 & 512 & +0.40 & -0.39 \\
    \midrule
    W4A16 & Prefill & 16 & 128   & 1   & -0.11 & +0.09 \\
    W4A16 & Prefill & 4  & 512   & 1   & +0.21 & -0.21 \\
    W4A16 & Prefill & 1  & 2,048 & 1   & +0.37 & -0.37 \\
    W4A16 & Prefill & 1  & 8,192 & 1   & +0.53 & -0.52 \\
    W4A16 & Decode  & 32 & 128   & 32  & +0.38 & -0.48 \\
    W4A16 & Decode  & 16 & 128   & 128 & +0.69 & -0.81 \\
    W4A16 & Decode  & 4  & 128   & 512 & +0.68 & -0.70 \\
    W4A16 & Decode  & 1  & 2,048 & 512 & +0.66 & -0.65 \\
    \bottomrule
  \end{tabular}%
  }
  \caption{Direct compact-serving comparison with uncompressed PLE for matched
  BF16 and W4A16 E4B checkpoints. Input and output lengths are per sequence;
  values are means over six timed measurements. Positive wall-time change
  means slower, and positive throughput change means faster.}
  \label{tab:appendix-runtime}
\end{table*}

Table~\ref{tab:appendix-runtime} lists the eight workloads; the fixed 1~GiB
cache isolates the cost of compact lookup. To measure the effect of the freed
memory in local use, we start both conditions with the same GPU memory
utilization, so that vLLM assigns all remaining memory to the key--value
cache, submit one to sixteen requests at once with 1,024 output tokens each,
and record the time until all of them finish. Prompts are random ordinary
token IDs, generation ignores end-of-sequence tokens, and each time is the
mean of two repeats after warm-up. Utilizations of 0.90, 0.66, 0.50 and 0.33
of the 24~GiB RTX~4090 give 21.6, 15.8, 12.0 and 7.9~GiB, written 21.6, 16,
12 and 8~GiB. For BF16 E2B we serve the default OPTIC
artifact of Table~\ref{tab:primary-matrices}; vLLM loads 9.89~GiB of weights
with dense PLE and 5.56~GiB with compact PLE, compared with 15.28 and
10.08~GiB for BF16 E4B. vLLM assigns 4.87 and 10.06~GiB to the key--value
cache for dense and compact BF16 E4B at 21.6~GiB and 4.42~GiB for compact at
16~GiB; for BF16 E2B it assigns 0.82 and 5.15~GiB at 12~GiB and 1.15~GiB for
compact at 8~GiB.
Table~\ref{tab:local-serving} reports the times. When all requests
fit in the cache, the conditions take nearly the same time; otherwise dense
PLE runs them in turns. W4A16 compact runs use the fused decode in the released
code. Dense W4A16 E4B fits within 16~GiB, where compact PLE finishes 8 and
16 requests 1.20$\times$ faster; at 21.6~GiB, where the dense cache also holds
all requests, compact PLE takes up to 1.7\% longer with W4A16 E4B and 0.6\%
longer with BF16 E2B. With 8K-token prompts, BF16 E4B times differ by at most 1\%
for one to sixteen requests. At 16~GiB for BF16 E4B and 8~GiB for BF16 E2B, dense PLE
leaves no memory for the cache and the engine does not start, whereas compact
PLE matches or exceeds dense PLE at the larger budget except for two or four
64K-token E4B requests. With one request at a time and 512 output tokens,
dense BF16 PLE at 21.6~GiB completes an 88,000-token prompt in 21~s, whereas
a 96,000-token prompt is accepted but not scheduled within 15~min; compact
PLE completes a 130,000-token prompt in 33~s, and at 16~GiB completes 80,000
tokens but not 96,000. A retrieval test checks long-context use: each prompt
places the sentence ``The special magic number for \emph{word} is
\emph{number}'', with a random word and seven-digit number, in WikiText-2
test text~\citep{merity2016pointer} at a depth of 0, 25, 50, 75 or 100\% and
asks for the number, with eight prompts per length and depth and greedy
decoding. On BF16 E4B at 21.6~GiB, dense and compact PLE answer all 240
prompts of 4,096 to 81,920 tokens correctly with identical outputs, and compact PLE
answers all 120 prompts of 98,304 to 130,000 tokens correctly.

\begin{table}[!ht]
  \centering
\resizebox{\linewidth}{!}{%
  \begin{tabular}{@{}rrrrrr@{}}
    \toprule
    & & \multicolumn{2}{c}{Same budget} & & Smaller \\
    \cmidrule(lr){3-4}
    Input & Requests & Dense & Compact & Speedup & Compact \\
    \midrule
    \multicolumn{6}{@{}l}{\textit{BF16 E4B: 21.6~GiB; smaller budget 16~GiB}} \\
    32K & 1 & 17.3 & 17.3 & 1.00$\times$ & 17.2 \\
    32K & 2 & 22.2 & 22.2 & 1.00$\times$ & 22.0 \\
    32K & 4 & 30.1 & 29.9 & 1.01$\times$ & 29.7 \\
    32K & 8 & 59.9 & 46.1 & 1.30$\times$ & 59.9 \\
    32K & 16 & 105.1 & 78.7 & 1.34$\times$ & 105.6 \\
    64K & 1 & 24.0 & 23.8 & 1.01$\times$ & 23.8 \\
    64K & 2 & 34.5 & 34.4 & 1.00$\times$ & 47.7 \\
    64K & 4 & 68.2 & 54.5 & 1.25$\times$ & 95.3 \\
    \midrule
    \multicolumn{6}{@{}l}{\textit{W4A16 E4B: 16~GiB}} \\
    32K & 1 & 11.9 & 12.0 & 0.99$\times$ & -- \\
    32K & 4 & 24.2 & 24.0 & 1.01$\times$ & -- \\
    32K & 8 & 48.3 & 40.4 & 1.20$\times$ & -- \\
    32K & 16 & 88.2 & 73.2 & 1.21$\times$ & -- \\
    \midrule
    \multicolumn{6}{@{}l}{\textit{BF16 E2B: 12~GiB; smaller budget 8~GiB}} \\
    8K & 1 & 7.6 & 7.6 & 1.00$\times$ & 7.6 \\
    8K & 4 & 9.6 & 9.7 & 1.00$\times$ & 9.6 \\
    8K & 16 & 23.0 & 15.4 & 1.49$\times$ & 15.3 \\
    32K & 1 & 10.8 & 10.8 & 1.00$\times$ & 10.7 \\
    32K & 2 & 13.9 & 14.0 & 1.00$\times$ & 13.8 \\
    32K & 4 & 27.8 & 19.1 & 1.45$\times$ & 18.9 \\
    32K & 8 & 55.7 & 30.1 & 1.85$\times$ & 37.7 \\
    32K & 16 & 111.4 & 52.6 & 2.12$\times$ & 75.4 \\
    \bottomrule
  \end{tabular}%
  }
  \caption{Seconds until all concurrent requests finish on one RTX~4090, with
  1,024 output tokens per request. Speedup compares the two conditions at the
  same budget and uses unrounded times; dense PLE does not start at the
  smaller budget.}
  \label{tab:local-serving}
\end{table}

Table~\ref{tab:appendix-host-lookup} times one PLE lookup outside vLLM when
the table stays in pinned host memory. Dense lookup gathers BF16 rows on the
host and copies them to the GPU; compact lookup gathers and copies the codes
and decodes them on the GPU into BF16 rows. The decode folds the bias
into one float16-input, float32-accumulation multiplication and replays batches of up to 512 tokens from a CUDA graph. Dense lookup is faster
for a single token, and compact lookup is faster from 32 tokens.
\begin{table}[!ht]
  \centering
\resizebox{0.78\linewidth}{!}{%
  \begin{tabular}{@{}rrrrr@{}}
    \toprule
    & \multicolumn{2}{c}{Dense} & Compact & \\
    \cmidrule(lr){2-3}\cmidrule(lr){4-4}
    Tokens & GPU & Host & Host & Speedup \\
    \midrule
    1 & 18 & 27 & 46 & 0.6$\times$ \\
    8 & 18 & 63 & 66 & 0.9$\times$ \\
    32 & 18 & 126 & 69 & 1.8$\times$ \\
    128 & 18 & 380 & 90 & 4.2$\times$ \\
    512 & 26 & 1,092 & 154 & 7.1$\times$ \\
    2,048 & 90 & 4,077 & 521 & 7.8$\times$ \\
    8,192 & 401 & 18,617 & 1,537 & 12.1$\times$ \\
    \bottomrule
  \end{tabular}%
  }
  \caption{Median time in microseconds to return the BF16 PLE rows of all
  42 layers for uniformly random token IDs on an RTX~4090. Speedup compares
  compact and dense lookup from host memory and uses unrounded medians.}
  \label{tab:appendix-host-lookup}
\end{table}

\section{Additional Results}
\label{app:results}

Retention values follow Equation~\ref{eq:appendix-seven-task}; resident bytes
include the full compact representation used by the decoder.
Tables~\ref{tab:appendix-bf16-per-task} and
\ref{tab:appendix-w4a16-per-task} report the seven constituent benchmark
scores at the nominal $100\times$ tier.

\begin{table*}[t]
  \centering
  \resizebox{0.99\textwidth}{!}{%
  \begin{tabular}{@{}lrrrrrrrrr@{}}
    \toprule
    Method & $R$ & PIQA & GSM8K & \shortstack{GPQA\\Diamond}
      & \shortstack{AIME\\2026} & IFBench & \shortstack{Infographic\\VQA}
      & \shortstack{MMLU-\\Redux} & Retention \\
    \midrule
    \multicolumn{10}{l}{\emph{E2B}} \\
    Uncompressed PLE & 1.00 & 72.4 & 90.5 & 43.2 & 42.5 & 36.0 & 63.1 & 71.5 & 100.0 \\
    \midrule
    \multicolumn{10}{l}{\emph{Baselines}} \\
    PQ & 101.24 & 68.3 & 53.1 & 29.8 & 0.8 & 0.7 & 19.2 & 55.7 & 22.9 \\
    RVQ & 100.96 & 70.7 & 39.7 & 23.0 & 0.0 & 1.0 & 23.6 & 58.9 & 0.0 \\
    OPQ-P & 100.92 & 69.2 & 51.5 & 1.5 & 0.0 & 1.3 & 42.1 & 56.4 & 0.0 \\
    AQ & 100.96 & 70.0 & 17.3 & 1.3 & 0.0 & 2.0 & 40.8 & 55.1 & 0.0 \\
    AQLM Euclidean beam encoder
      & 100.96 & 70.3 & 45.7 & 27.0 & 1.7 & 0.3 & 11.0 & 60.5 & 20.7 \\
    INT8 randomized low-rank SVD
      & 100.13 & 71.2 & 28.1 & 0.4 & 0.0 & 6.7 & 45.4 & 52.4 & 0.0 \\
    \midrule
    \multicolumn{10}{l}{\emph{Our methods}} \\
    \Owpq{}
      & 96.14 & 71.4 & 85.4 & 29.5 & 8.3 & 0.7 & 11.2 & 62.9 & 32.2 \\
    OPTIC
      & 99.86 & 72.0 & 87.4 & 32.2 & 32.5 & 38.0 & 58.5 & 62.5 & 89.1 \\
    Frequency-tiered OPTIC
      & 100.48 & 71.9 & 86.7 & 32.1 & 35.0 & 34.3 & 62.0 & 62.8 & \textbf{90.0} \\
    \midrule
    \multicolumn{10}{l}{\emph{E4B}} \\
    Uncompressed PLE & 1.00 & 75.9 & 92.9 & 56.9 & 45.8 & 44.0 & 69.7 & 78.9 & 100.0 \\
    \midrule
    \multicolumn{10}{l}{\emph{Baselines}} \\
    PQ & 101.43 & 75.4 & 73.5 & 47.5 & 20.8 & 16.0 & 38.4 & 74.5 & 66.3 \\
    RVQ & 98.64 & 75.8 & 70.5 & 47.5 & 6.7 & 1.0 & 37.4 & 75.2 & 37.6 \\
    OPQ-P & 101.16 & 75.6 & 67.7 & 47.6 & 14.2 & 6.7 & 28.4 & 73.8 & 52.4 \\
    AQ & 101.20 & 76.1 & 70.7 & 49.0 & 4.2 & 5.3 & 41.9 & 74.8 & 45.6 \\
    AQLM Euclidean beam encoder
      & 101.20 & 76.4 & 74.1 & 46.1 & 16.7 & 2.0 & 33.6 & 74.7 & 46.7 \\
    INT8 randomized low-rank SVD
      & 99.76 & 76.1 & 51.4 & 46.6 & 3.3 & 2.7 & 40.3 & 73.7 & 37.7 \\
    \midrule
    \multicolumn{10}{l}{\emph{Our methods}} \\
    \Owpq{}
      & 96.36 & 75.9 & 93.3 & 54.5 & 49.2 & 33.0 & 61.8 & 76.0 & 93.3 \\
    OPTIC
      & 99.94 & 75.9 & 92.8 & 54.7 & 48.3 & 41.7 & 67.5 & 76.6 & \textbf{97.8} \\
    Frequency-tiered OPTIC
      & 100.00 & 75.9 & 92.8 & 53.7 & 40.8 & 40.0 & 69.5 & 76.4 & 95.7 \\
    \bottomrule
  \end{tabular}%
  }
  \caption{Per-task results (\%) for the original BF16 releases at the nominal
  $100\times$ tier. $R$ is the actual resident-byte PLE compression ratio.
  Bold marks the highest compressed-model retention for each model size.}
  \label{tab:appendix-bf16-per-task}
\end{table*}

\begin{table*}[t]
  \centering
  \resizebox{0.99\textwidth}{!}{%
  \begin{tabular}{@{}lrrrrrrrrr@{}}
    \toprule
    Method & $R$ & PIQA & GSM8K & \shortstack{GPQA\\Diamond}
      & \shortstack{AIME\\2026} & IFBench & \shortstack{Infographic\\VQA}
      & \shortstack{MMLU-\\Redux} & Retention \\
    \midrule
    \multicolumn{10}{l}{\emph{E2B}} \\
    Uncompressed PLE & 1.00 & 74.7 & 88.1 & 41.8 & 25.0 & 33.7 & 56.0 & 68.1 & 100.0 \\
    \midrule
    \multicolumn{10}{l}{\emph{Baselines}} \\
    PQ & 101.24 & 71.3 & 42.5 & 26.0 & 0.0 & 23.0 & 20.8 & 53.9 & 0.0 \\
    RVQ & 100.96 & 72.5 & 51.9 & 22.9 & 0.8 & 6.3 & 6.4 & 56.2 & 29.3 \\
    OPQ-P & 100.92 & 71.6 & 52.7 & 23.6 & 0.0 & 5.3 & 11.5 & 55.3 & 0.0 \\
    AQ & 100.96 & 70.9 & 13.0 & 8.1 & 0.0 & 1.3 & 25.5 & 49.9 & 0.0 \\
    AQLM Euclidean beam encoder
      & 100.96 & 72.7 & 55.3 & 28.3 & 0.8 & 11.7 & 4.9 & 56.8 & 32.0 \\
    INT8 randomized low-rank SVD
      & 100.13 & 72.9 & 26.8 & 25.9 & 0.0 & 8.0 & 29.4 & 52.0 & 0.0 \\
    \midrule
    \multicolumn{10}{l}{\emph{Our methods}} \\
    \Owpq{}
      & 96.14 & 72.4 & 82.2 & 27.4 & 13.3 & 3.7 & 6.5 & 59.6 & 44.5 \\
    OPTIC
      & 99.86 & 73.6 & 84.2 & 30.7 & 30.0 & 35.3 & 48.3 & 61.7 & 91.5 \\
    Frequency-tiered OPTIC
      & 100.48 & 73.0 & 84.0 & 31.7 & 25.0 & 34.7 & 54.6 & 59.5 & \textbf{93.0} \\
    \midrule
    \multicolumn{10}{l}{\emph{E4B}} \\
    Uncompressed PLE & 1.00 & 77.7 & 93.6 & 51.1 & 39.2 & 40.0 & 65.5 & 77.7 & 100.0 \\
    \midrule
    \multicolumn{10}{l}{\emph{Baselines}} \\
    PQ & 101.43 & 76.4 & 79.7 & 45.8 & 20.8 & 34.3 & 52.1 & 73.7 & 82.4 \\
    RVQ & 101.20 & 76.8 & 78.7 & 44.8 & 18.3 & 41.7 & 50.6 & 73.5 & 82.0 \\
    OPQ-P & 101.16 & 76.9 & 69.8 & 45.1 & 7.5 & 37.0 & 13.4 & 72.0 & 57.9 \\
    AQ & 101.20 & 76.1 & 61.6 & 42.7 & 4.2 & 35.0 & 54.0 & 72.4 & 62.8 \\
    AQLM Euclidean beam encoder
      & 101.20 & 76.8 & 76.3 & 45.6 & 20.0 & 38.7 & 51.6 & 74.4 & 82.8 \\
    INT8 randomized low-rank SVD
      & 100.24 & 75.8 & 49.3 & 39.8 & 3.3 & 18.7 & 40.2 & 72.3 & 51.1 \\
    \midrule
    \multicolumn{10}{l}{\emph{Our methods}} \\
    \Owpq{}
      & 96.36 & 76.4 & 92.7 & 50.9 & 35.8 & 36.7 & 34.9 & 74.8 & 88.3 \\
    OPTIC
      & 99.94 & 76.2 & 92.9 & 48.0 & 41.7 & 41.7 & 59.2 & 75.3 & \textbf{96.9} \\
    Frequency-tiered OPTIC
      & 100.00 & 76.9 & 93.1 & 48.1 & 35.0 & 37.0 & 61.3 & 76.1 & 95.1 \\
    \bottomrule
  \end{tabular}%
  }
  \caption{Per-task results (\%) for the official W4A16 releases at the nominal
  $100\times$ tier. Notation follows
  Table~\ref{tab:appendix-bf16-per-task}. Bold marks the highest compressed-model
  retention for each model size.}
  \label{tab:appendix-w4a16-per-task}
\end{table*}

\subsection{Original BF16}

Table~\ref{tab:primary-matrices} reports the nominal-$100\times$ comparison on
both model scales. Table~\ref{tab:appendix-cross-rate-bf16-e4b} extends the E4B
comparison across the practical compression range.

\begin{table*}[t]
  \centering
\resizebox{0.85\textwidth}{!}{%
  \begin{tabular}{@{}lrrrrrr@{}}
    \toprule
    & \multicolumn{2}{c}{$20\times$} & \multicolumn{2}{c}{$40\times$}
      & \multicolumn{2}{c}{$100\times$} \\
    \cmidrule(lr){2-3}\cmidrule(lr){4-5}\cmidrule(lr){6-7}
    Method & Actual $R$ & Retention & Actual $R$ & Retention
      & Actual $R$ & Retention \\
    \midrule
    \multicolumn{7}{l}{\emph{Baselines}} \\
    PQ & 21.17 & \textbf{98.6} & 42.00 & 77.0 & 101.43 & 66.3 \\
    RVQ & 20.26 & 94.2 & 40.52 & 86.0 & 98.64 & 37.6 \\
    OPQ-P & 21.15 & 98.1 & 41.95 & 90.4 & 101.16 & 52.4 \\
    AQLM Euclidean beam encoder & 20.24 & 98.2 & 42.16 & 82.5 & 101.20 & 46.7 \\
    INT8 randomized low-rank SVD & 20.01 & 68.5 & 40.02 & 52.7 & 99.76 & 37.7 \\
    \midrule
    \multicolumn{7}{l}{\emph{Our methods}} \\
    \Owpq{} & 20.31 & 98.3 & 40.29 & 90.8 & 96.36 & 93.3 \\
    OPTIC & 19.99 & 97.4 & 40.04 & 96.4 & 99.94 & \textbf{97.8} \\
    Frequency-tiered OPTIC & 20.09 & 94.9 & 40.03 & \textbf{97.0} & 100.00 & 95.7 \\
    \bottomrule
  \end{tabular}%
  }
  \caption{Original BF16 E4B at the nominal $20\times$, $40\times$, and
  $100\times$ PLE compression tiers. $R$ is the actual resident-byte
  compression ratio; retention (\%) is the seven-task geometric mean. Bold
  marks the highest retention at each tier.}
  \label{tab:appendix-cross-rate-bf16-e4b}
\end{table*}

\subsection{Official W4A16}

Table~\ref{tab:appendix-w4a16-practical-range} compares three baselines with
the two OPTIC variants. The full
nominal-$100\times$ comparison, including AQ and the adapted AQLM and SVD
baselines, appears in Table~\ref{tab:primary-matrices}.

\begin{table*}[t]
  \centering
\resizebox{0.88\textwidth}{!}{%
  \begin{tabular}{@{}lrrrrrr@{}}
    \toprule
    & \multicolumn{3}{c}{E2B W4A16} & \multicolumn{3}{c}{E4B W4A16} \\
    \cmidrule(lr){2-4}\cmidrule(lr){5-7}
    Method & $20\times$ & $40\times$ & $100\times$
           & $20\times$ & $40\times$ & $100\times$ \\
    \midrule
    \multicolumn{7}{l}{\emph{Baselines}} \\
    PQ & 21.13/\textbf{95.0} & 41.87/85.6 & 101.24/0.0
       & 21.17/96.0 & 42.00/88.6 & 101.43/82.4 \\
    RVQ & 20.17/85.4 & 40.35/41.4 & 100.96/29.3
        & 20.26/96.5 & 40.52/89.3 & 101.20/82.0 \\
    OPQ-P & 21.12/94.9 & 41.81/83.7 & 100.92/0.0
          & 21.15/93.6 & 41.95/91.6 & 101.16/57.9 \\
    \midrule
    \multicolumn{7}{l}{\emph{Our methods}} \\
    OPTIC
      & 20.01/93.9 & 39.95/\textbf{93.8} & 99.86/91.5
      & 19.99/96.8 & 40.04/95.9 & 99.94/\textbf{96.9} \\
    Frequency-tiered OPTIC
      & 20.06/93.8 & 40.19/93.5 & 100.48/\textbf{93.0}
      & 20.09/\textbf{97.7} & 40.03/\textbf{96.5} & 100.00/95.1 \\
    \bottomrule
  \end{tabular}%
  }
  \caption{Official W4A16 results at the nominal $20\times$, $40\times$, and
  $100\times$ PLE compression tiers. Each cell reports actual
  $R$/seven-task retention (\%) relative to the matched uncompressed PLE.
  Bold marks the highest retention for each model and tier.}
  \label{tab:appendix-w4a16-practical-range}
\end{table*}

\begin{table*}[t]
  \centering
\resizebox{0.7\textwidth}{!}{%
  \begin{tabular}{@{}ccccc@{}}
    \toprule
    Nominal tier & Actual $R$ & Resident PLE state
      & BF16 retention & W4A16 retention \\
    \midrule
    $20\times$   & 19.994  & 281.9~MB & 97.4 & 96.8 \\
    $40\times$   & 40.035  & 140.8~MB & 96.4 & 95.9 \\
    $100\times$  & 99.941  &  56.4~MB & 97.8 & 96.9 \\
    $500\times$  & 594.431 &   9.5~MB & 90.5 & 87.7 \\
    $1000\times$ & 1002.189 &  5.6~MB & 84.1 & 81.8 \\
    \bottomrule
  \end{tabular}%
  }
  \caption{E4B OPTIC across weight formats. Retention (\%) is
  the seven-task geometric mean; nominal labels denote target compression
  tiers, while $R$ reports the measured resident-byte ratio.}
  \label{tab:appendix-e4b-extreme-rate}
\end{table*}

\subsection{Sampling Uncertainty}
We form 95\% percentile intervals from 10,000 paired bootstrap
resamples~\citep{efron2000introduction} using seed 20260906. Resampling is paired
within each benchmark; GPQA permutations and AIME attempts are grouped by
their original question. Every resample recomputes capped reference-normalized
retention and the equal-weight seven-task geometric mean. The intervals reflect
benchmark-item sampling uncertainty for the fixed model, artifact, and
evaluation configuration.

\section{Output-Path-Weighted PCA-PQ}
\label{app:pca-pq}
Standard product quantization (PQ) stores one code tuple for each token--layer
row and fits its codebooks under Euclidean reconstruction error.
Output-path-weighted PCA-PQ retains this row-wise representation but derives
its distortion geometry from the local PLE projection defined in
Equation~\ref{eq:ple-path}. This complements OPTIC: the latter
changes the code topology by sharing one tuple across layers, whereas this
method retains one tuple per $(t,\ell)$ and changes the reconstruction metric.

For coordinate $j\in\{1,\ldots,d\}$, define its marginal sensitivity
\begin{equation}
\begin{aligned}
\omega_{\ell j}
 &=\frac{\tau^2}{n}\sum_{i\in\mathcal I_{\mathrm{fit}}}
       g_{i\ell j}^2\|\mathbf W^{\mathrm{out}}_{\ell,:,j}\|_2^2,\\
\overline{\omega}
 &=\frac{1}{Ld}\sum_{\ell=1}^{L}\sum_{j=1}^{d}\omega_{\ell j}.
\end{aligned}
  \label{eq:pca-pq-sensitivity}
\end{equation}
Let $\epsilon_w>0$ be a sensitivity floor and
$\alpha_w\in(0,1]$ a tempering exponent. We define
\begin{equation}
\begin{aligned}
a_{\ell j}
 &=\left[\max\left(\frac{\omega_{\ell j}}{\overline{\omega}},
                             \epsilon_w\right)\right]^{\alpha_w},\\
s_{\ell j}^2
 &=\frac{a_{\ell j}}{\frac{1}{Ld}\sum_{\ell'=1}^{L}
                         \sum_{j'=1}^{d}a_{\ell'j'}},\\
\mathbf D_\ell&=\operatorname{Diag}(\mathbf s_\ell).
\end{aligned}
  \label{eq:pca-pq-coordinate-scale}
\end{equation}
Thus, Euclidean error after multiplication by $\mathbf D_\ell$ corresponds to a power-tempered diagonal sensitivity surrogate.

Let
$\boldsymbol\mu=(VL)^{-1}\sum_{t=1}^{V}\sum_{\ell=1}^{L}
\mathbf D_\ell\mathbf e_{t\ell}\in\mathbb R^d$
be the mean of all scaled rows. We fit one global orthogonal PCA basis
$\mathbf R\in\mathbb R^{d\times d}$, with
$\mathbf R^{\mathsf T}\mathbf R=\mathbf I_d$, and encode
\begin{equation}
\begin{aligned}
\mathbf u_{t\ell}
 &=\mathbf R^{\mathsf T}(\mathbf D_\ell\mathbf e_{t\ell}
                            -\boldsymbol\mu)\in\mathbb R^d,\\
\widetilde{\mathbf u}_{t\ell}
 &=\operatorname{PQ}_{\boldsymbol\theta}(\mathbf u_{t\ell})
      \in\mathbb R^d.
\end{aligned}
  \label{eq:pca-pq-encode}
\end{equation}
Here $\widetilde{\mathbf u}_{t\ell}$ concatenates the codewords selected by
the row's code tuple from a single PQ bank shared across all $VL$ rows.

The inverse transform first produces
$\mathbf e^{(0)}_{t\ell}\in\mathbb R^d$. Let
$\mathcal T_{\mathrm{res}}\subseteq\{1,\ldots,V\}$ be the tokens with stored
residual corrections,
$\boldsymbol\rho_{t\ell}\in\mathbb R^d$ the stored correction for
$t\in\mathcal T_{\mathrm{res}}$, and
$\boldsymbol\gamma_\ell\in\mathbb R_{>0}^{d}$ the stored layer-coordinate
post-quantization scale. The requested-row decoder is
\begin{align}
  \mathbf e^{(0)}_{t\ell}
  &=
  \mathbf D_\ell^{-1}
  \left(
    \mathbf R\widetilde{\mathbf u}_{t\ell}
    +\boldsymbol\mu
  \right), \nonumber\\
  \mathbf e^{(1)}_{t\ell}
  &=
  \mathbf e^{(0)}_{t\ell}
  +
  \mathbf 1\{t\in\mathcal T_{\mathrm{res}}\}
  \boldsymbol\rho_{t\ell}, \nonumber\\
  \operatorname{Dec}_{\boldsymbol\theta}(t,\ell)
  &=
  \boldsymbol\gamma_\ell\odot\mathbf e^{(1)}_{t\ell}.
  \label{eq:pca-pq-decode}
\end{align}

\begin{algorithm}[H]
  \caption{Fitting procedure for output-path-weighted PCA-PQ.}
  \label{alg:pca-pq}
  \begin{algorithmic}[1]\small
    \RequireText{$\mathbf E$, $\mathcal I_{\mathrm{fit}}$, frozen gates and
      projections; a fixed PQ configuration and residual budget}
    \Ensure Compact representation $\boldsymbol\theta$
    \StateText{Compute $\{\omega_{\ell j}\}$ and $\{\mathbf D_\ell\}$ using
      Equations~\ref{eq:pca-pq-sensitivity}--%
      \ref{eq:pca-pq-coordinate-scale}}
    \StateText{Fit $\boldsymbol\mu$ and the global PCA basis $\mathbf R$ to the
      scaled token--layer rows}
    \StateText{Fit one shared PQ bank and the row assignments to
      $\{\mathbf u_{t\ell}\}_{t,\ell}$}
    \StateText{Select $\mathcal T_{\mathrm{res}}$ using calibration-token counts
      and store the residual corrections $\{\boldsymbol\rho_{t\ell}\}$}
    \StateText{Fit $\{\boldsymbol\gamma_\ell\}$ to the residual-corrected
      reconstruction}
    \State \Return $\boldsymbol\theta$
  \end{algorithmic}
\end{algorithm}

\section{Controlled Comparisons}
\label{app:controlled-fitting}
In the code-sharing comparison of Section~\ref{sec:ablations}, independent
latent codes use the main method's output-path-weighted initialization and
unused-codeword handling. The objective and optimization comparisons instead use Euclidean PCA and
$k$-means initialization without post-initialization unused-codeword handling, so
that every objective starts from the same codes.
Table~\ref{tab:main-comparison} reports the representation comparison under
this common initialization over fit seeds 0, 1 and 2.

\begin{table}[!t]
\centering
\resizebox{0.97\linewidth}{!}{%
\begin{tabular}{lrr}
\toprule
Representation & E4B BF16 & E4B W4A16 \\
\midrule
Independent latent codes & $75.90\pm3.38$ & $81.51\pm2.06$ \\
Additive codes & $82.66\pm5.06$ & $90.95\pm1.23$ \\
OPTIC & $95.49\pm0.40$ & $95.16\pm0.42$ \\
\bottomrule
\end{tabular}%
}
\caption{Seven-task retention (\%) near $100\times$ PLE compression. Each cell reports the mean and sample standard deviation over fit seeds 0, 1, and 2. All rows use output-path fitting and a common Euclidean initialization policy. Table~\ref{tab:sharing-ablation} reports the seed-0 comparison under the main fitting recipe.}
\label{tab:main-comparison}
\end{table}

The E4B shared representation uses 161 groups of width four, 256 codewords
per group and 56,404,992 resident bytes. Independent latent coding uses
five groups of width four per layer and 55,931,904 bytes. Additive coding
uses five codebooks and 55,705,600 bytes. Each representation uses the
same mixed calibration policy and 5,000 continuous updates. Alternating
fits perform two reassignment sweeps after each 500-update block through
update 4500. The fitting seeds are 0, 1 and 2.

For E4B, the calibration population contains 786,432 natural positions
from 1,536 sequences of length 512, 262,144 vocabulary rows and 55,200
repeated special-token rows. Natural contexts come from FineWeb,
FineWeb-Edu, OpenWebMath and FineWeb2. Synthetic rows use all-ones gates.
Each checkpoint or QAT parent supplies its own gates and projections.

\subsection{Objective and Optimization}
The objective comparison holds initial codes and continuous parameters
fixed across Euclidean, diagonal output-path and full output-path losses.
It uses 5,000 continuous updates. The Euclidean loss is about $10^4$
times smaller than the output-path losses. Each loss is divided by its
value for an all-zero reconstruction, with the same learning rate and
gradient-clipping threshold across objectives.
Table~\ref{tab:objective} reports all three fitting seeds for each loss
through their means and sample standard deviations. The paired mean gain
of the full objective over Euclidean fitting is $1.98\pm1.19$ percentage
points; its gain over the diagonal approximation is $0.31\pm0.45$ points.

\label{app:optimization}
The optimization comparison uses initialization alone, continuous fitting
with codes fixed, or continuous fitting with periodic reassignment.
The latter two both take 5,000 continuous updates; reassignment adds
nine rounds of two sweeps. Their mean retention difference is 0.83
percentage points, with paired gains from 0.65 to 1.14 points across
seeds. Initialization performs no continuous update.

\subsection{Evaluation Details}
The controlled comparisons use checkpoint-matched uncompressed evaluations
from the same task runners. Fitting seeds index independent compact fits;
benchmark attempt seeds and question identities follow the fixed protocols.
The following descriptions specify the prompts and scoring used for
all evaluations in this paper.
\paragraph{PIQA.}
We score all 1,838 validation items with zero demonstrations.
The score is accuracy under raw choice log likelihood, without
length normalization or answer generation. The context is
\texttt{Question: \{goal\}}, a newline, then \texttt{Answer:}.
Each solution is appended directly, with no added separator.
The tokenizer encodes the context and complete context--solution
string without special tokens; one beginning-of-sequence token is
prepended. Scoring sums actual-token log probabilities from
the context token count plus one through the end of the complete
sequence. The larger sum wins; ties select the first solution.

\paragraph{GSM8K.}
We evaluate all 1,319 test questions with eight fixed demonstrations
from the training split. Generation is greedy, thinking is disabled,
and the maximum number of new tokens is 8,192. The scorer uses EvalScope answer extraction and mathematical equivalence.
The same eight examples are used for every method.

\paragraph{GPQA Diamond.}
The population is 198 questions with four fixed answer permutations,
giving 792 scored attempts. Accuracy is the micro-average over attempts,
over all four permutations. We use zero demonstrations, thinking enabled,
temperature 1.0, top-$p$ 0.95, top-$k$ 64, and a maximum of 16,384 new tokens.
The prompt asks for a final line of the form
\texttt{Answer: A}, with the choices labeled A--D.
The strict final-letter parser treats extraction failures as wrong.
The dataset revision is \texttt{633f5ee89ab8}.

\paragraph{AIME 2026.}
We evaluate 30 questions with attempt seeds 0--3, for 120 attempts.
The score is micro accuracy over all attempts. The zero-shot prompt requests an integer between 0 and 999
in a boxed final answer. Thinking is enabled, temperature is 1.0,
top-$p$ is 0.95, top-$k$ is 64 and the maximum number of new tokens is 32,768.
The MathArena boxed-integer rule supplies scoring.
Each of the 120 attempts contributes equally to accuracy.

\paragraph{IFBench.}
All 300 prompts are evaluated with zero demonstrations, thinking enabled,
temperature 0, top-$p$ 1.0 and a maximum of 16,384 new tokens.
The metric is prompt-level loose accuracy from the author verifier,
revision \texttt{db69a6f05689}~\citep{pyatkin2026generalizing}.
Empty and truncated outputs remain scored. The complete generated answer is passed to the verifier.

\paragraph{InfographicVQA.}
We use all 2,801 validation questions and original ANLS scoring.
Images precede text, with up to 1,120 visual soft tokens per image.
The question is followed by an instruction to answer using one word
or phrase. There are no demonstrations. Thinking is enabled,
temperature is 1.0, top-$p$ is 0.95, top-$k$ is 64, and the maximum number
of new tokens is 4,096. The prompt follows the lmms-eval adaptation, with the same input construction for all compared methods.

\paragraph{MMLU-Redux.}
We use 100 questions from each of 30 fixed subjects, totaling 3,000 items,
from revision \texttt{3720db6aeb3d}~\citep{hendrycks2020measuring,gema2025we}.
The prompt asks which option A--D is correct for the named category
and requests a final phrase of the form
\texttt{The best answer is (A)}.
We use zero demonstrations, thinking disabled, temperature 1.0,
top-$p$ 0.95, top-$k$ 64 and a maximum of 8,192 new tokens.
The last matching answer is scored; a parsing failure is wrong.

\paragraph{Belebele.}
Outside the seven tasks, we evaluate BF16 E4B on the 900 questions of
Belebele~\citep{bandarkar2024belebele} in English and in Simplified
Chinese, with zero demonstrations, an English instruction and thinking
disabled. The prediction is the answer label A--D with the highest log
probability. At about $100\times$ PLE compression, OPTIC answers 92.0\% and
89.1\% of the questions correctly,
compared with 90.9\% and 89.8\% with uncompressed PLE.

\subsection{PLE Structure and Decoder-Rank Control}
\label{app:cross-layer-structure}
\paragraph{Decomposition.}
We use all 262,144 token rows and 42 layers of the original BF16 E4B PLE
table in float64, with every row weighted equally. The layer mean is
$\mathbf m_\ell=V^{-1}\sum_t\mathbf e_{t\ell}$ and the centered remainder
is $\mathbf r_{t\ell}=\mathbf e_{t\ell}-\mathbf m_\ell$. The layer means
hold $\sum_\ell V\|\mathbf m_\ell\|^2/\sum_{t,\ell}\|\mathbf e_{t\ell}\|^2
=8.10\%$ of the squared Frobenius norm. Their share within a single layer
ranges from 0.2\% to 33.7\%. Fitting initializes $\mathbf b_\ell$ at
$\mathbf m_\ell$. For the main BF16 E4B fit, with the biases and the means
stacked over layers into $42\times256$ matrices $\mathbf B$ and
$\mathbf M$, $\|\mathbf B-\mathbf M\|_F/\|\mathbf M\|_F=14.3\%$.

\paragraph{Spectra.}
For each layer we compute the eigenvalues of the covariance of
$\mathbf r_{t\ell}$ in the stored coordinates and of
$\overline{\mathbf G}_\ell^{1/2}\mathbf r_{t\ell}$ at the projection
output, where $\overline{\mathbf G}_\ell$ is
Equation~\ref{eq:latent-vq-marginal-metric} computed over the training
records of the main BF16 E4B fit, as in OPTIC's initialization. In the
stored coordinates, the 20 leading eigenvalues hold
10.5--24.9\% of a layer's trace and 90\% needs 212--227 eigenvalues; at
the projection output, they hold 25.0--98.2\% (median 45.0\%) and 90\%
needs 1--205 eigenvalues (median 158).

\paragraph{Layer-mean table.}
Replacing every vector with its layer mean and evaluating with the protocol
of Section~\ref{sec:experiments} gives PIQA 75.5\%, GSM8K 43.5\%, GPQA
Diamond 23.0\%, AIME~2026 0.0\%, IFBench 6.7\%, InfographicVQA 49.3\% and
MMLU-Redux 67.4\%.

\paragraph{Decoder-rank control.}
For each of the three E4B BF16 fits of Table~\ref{tab:main-comparison},
we keep the indices, codebooks and biases and
refit each layer's decoder weights $\mathbf A_\ell$ by minimizing
$\sum_t f_t\|\mathbf A_\ell\mathbf z_t+\mathbf b_\ell-\mathbf e_{t\ell}\|^2$
with $\operatorname{rank}(\mathbf A_\ell)\le r$, where $f_t$ counts token
$t$ in the training records of the fit. The solution truncates the
singular value decomposition of the whitened response in float64. We store
the result in float16, at the same 56,404,992 bytes. Rank 256 places no
constraint on a $256\times644$ matrix.

\end{document}